\documentclass[twocolumn,preprintnumbers,amsmath,amssymb]{revtex4}

\usepackage{graphicx}
\usepackage{dcolumn}
\usepackage{bm}
\usepackage{epstopdf}
\usepackage{natbib}

\bibpunct{[}{]}{;}{a}{}{,} 

\begin{document}


\title{Switcher-random-walks: a cognitive-inspired mechanism for network exploration}

\author{Joaqu\'in Go\~ni$^{1,2}$, I\~nigo Martincorena$^2$, Bernat Corominas-Murtra$^3$, Gonzalo Arrondo$^2$, Sergio Ardanza-Trevijano$^1$, Pablo Villoslada$^{2,4}$}

\affiliation{
$^1$ Department of Physics and Applied Mathematics. University of Navarra, Pamplona, Spain\\
$^2$ Department of Neurosciences. Center for Applied Medical Research. University of Navarra, Pamplona, Spain\\
$^3$ ICREA-Complex Systems Lab, Universitat Pompeu Fabra - Parc de Recerca Biom\`edica de Barcelona, Dr. Aiguader 80, 08003 Barcelona, Spain\\
$^4$ Department of Neurosciences. Institut d'investigacions Biom\`ediques August Pi i Sunyer (IDIBAPS). Hospital Cl\'inic, Barcelona, Spain.\\
}

\thanks{Corresponding AUTHOR: pvilloslada@clinic.ub.es\\}

\begin{abstract}

Semantic memory is the subsystem of human memory that stores knowledge
of concepts or meanings, as  opposed to life specific experiences. The
organization of concepts within semantic memory can be understood as a
semantic network,  where the concepts (nodes)  are associated (linked)
to others depending on perceptions, similarities, etc.  Lexical access
is the complementary  part of this system and  allows the retrieval of
such  organized knowledge.  While  conceptual information  is stored
under  certain  underlying organization  (and  thus  gives  rise to  a
specific topology), it is crucial to have an accurate access to any of
the information units, e.g.   the concepts, for efficiently retrieving
semantic  information  for  real-time  needings.   An  example  of  an
information retrieval  process occurs in verbal fluency  tasks, and it
is  known  to  involve  two  different  mechanisms:  `clustering',  or
generating  words within  a subcategory,  and, when  a  subcategory is
exhausted, `switching' to a new subcategory. We extended this approach
to random-walking on a  network (clustering) in combination to jumping
(switching)  to any  node  with certain  probability  and derived  its
analytical expression  based on Markov chains. Results  show that this
dual mechanism  contributes to  optimize the exploration  of different
network models in terms of  the mean first passage time. Additionally,
this cognitive inspired dual mechanism opens a new framework to better
understand  and   evaluate  exploration,  propagation   and  transport
phenomena in other complex  systems where switching-like phenomena are
feasible.

\bigskip


\end{abstract}

\maketitle

\section{Introduction}

Semantic memory  is a distinct  part of the declarative  memory system
\citep*{tulving1972}  comprising knowledge  of facts,  vocabulary, and
concepts acquired through  everyday life \citep*{squire1987}. Contrary
to episodic memory, which  stores life experiences, semantic memory is
not  linked to  any particular  time or  place. In  a  more restricted
definition, it  is responsible for the storage  of semantic categories
and naming of natural  and artificial concepts \citep*{budson2005}. It
is  known that  this memory  involves distinct  brain regions  and its
impairment  in  neurodegenerative  diseases  such  as  fronto-temporal
dementia \citep*{libon2007}, multiple sclerosis \citep*{henry2006} and
Alzheimer's   disease  \citep*{rogers2008}   produce   verbal  fluency
deficits.   For   this   reason,   lexical   access,   the   cognitive
information-retrieval  process in charge  of retrieving  concepts, has
been  widely explored  through semantic  verbal fluency  tasks  in the
context  of  neuropsychological  evaluation \citep*{lezak1995}.  These
tests  require the  generation of  words corresponding  to  a specific
semantic  category, typically animals,  fruits or  tools, for  a given
time. Although the  task is easy to explain, it  actually results in a
complex challenge where  retrieving as many concepts as  possible in a
limited  time  depends  more  on  cognitive  mechanisms  than  on  the
knowledge itself.  According to the two-component model proposed by A.
Troyer  \citep*{troyer1997}, optimal  fluency  performance involves  a
balance between  two different processes:  `clustering', or generating
words  within a  subcategory, and,  when a  subcategory  is exhausted,
`switching'  to a  new subcategory.   In the  case of  naming animals,
clustering    produces   semantically   related    transitions   (e.g.
\textit{lion-tiger}) and switching is  a mechanism that allows to jump
or  shift to different  semantic fields  (e.g.  \textit{tiger-shark}).
While the  former is attached to  the temporal lobe of  the brain, the
latter   has   been   associated    to   a   frontal   lobe   activity
\citep*{troyer2002}.  Evidences  of the interaction  between these two
regions of the brain during language related tasks has led a number of
studies   to  refer  to   a  \textit{fronto-temporal   modulation}  or
interaction \citep*{poldrack1999,troyer2002}.

In  this paper,  the cognitive  paradigm that  consists  of retrieving
words   from   a   \textit{semantic   network}   \citep*{thornton2002,
  rogers2008}  was generalized to  an exploration  task on  a network.
Clustering was modeled as  a random-walker constrained to the topology
of the network and switching as an extra-topological mechanism that is
able   to   move   from   any   node   to   any   node   (see   figure
\ref{figSRWexample}). The combination of these two processes gave rise
to  a dual mechanism  denoted here  as \textit{switcher-random-walker}
(SRW), i.e.  a random-walker with the additional ability of switching.
The combination  of switching and  clustering, i.e., free  jumping and
random walking, was ruled by a parameter $q$, which is the probability
of  switching  at   every  step,  and  thus  is   the  parameter  that
metaphorically  rules the  fronto-temporal  modulation. Therefore  the
complementary $(1-q)$ is the  probability of clustering at every step,
and  can   be  interpreted  as  the  strength   of  the  \textit{local
  perseverance} of the exploration before moving somewhere else within
the network (specially for  those networks with either high clustering
coefficient  or high  modularity).  This  cognitive  inspired paradigm
gives  rise to  the following  question:  how does  switching and  its
modulation affect random exploration of different network models?
 

\begin{figure}
\begin{center}
\includegraphics[width=8cm]{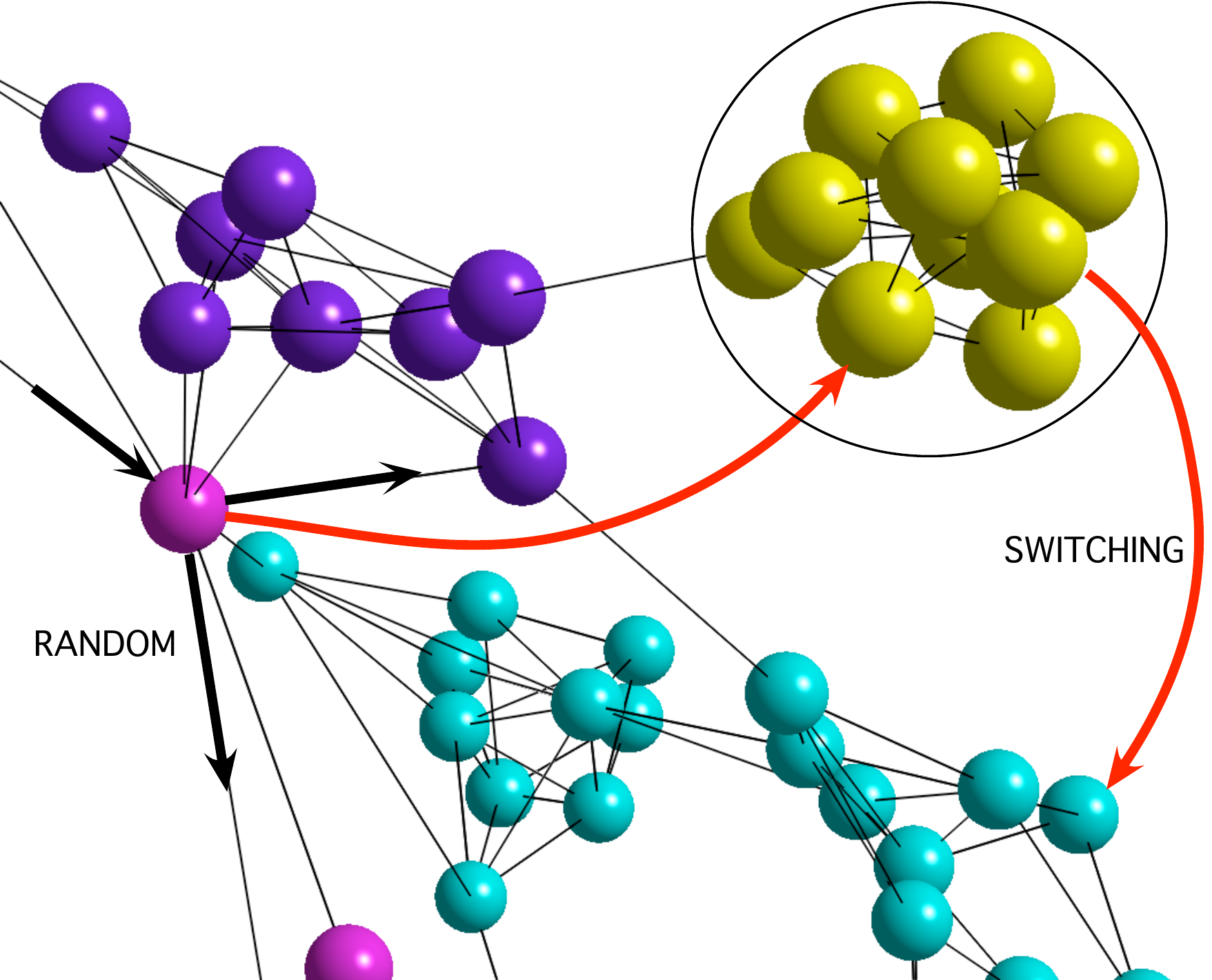}
\end{center}
\caption{\footnotesize{  Switcher  random  walks: transitions  between
    nodes in a graph can  occur through random movements following the
    edges  (black  arrows) but  also  through  switches (red  arrows).
    Switching  allows a  more efficient  exploration,  since clustered
    graphs    might    difficult    finding   rare    paths    between
    sub-graphs. Isolated modules in particular (circle) would be seldom
    reached and rarely abandoned through random walking.  \hfill}}
\label{figSRWexample}
\end{figure}
Search,  propagation  and transport  phenomena  have  been studied  in
networks \citep*{bollt2005}, where it is crucial to define whether the
full topology is  known. When it is known, the ease  to reach any node
from  any  node  is  measured  by the  \textit{shortest  path  length}
\citep*{watts1998,tadic2002}. When it  remains unknown, exploration is
modeled by  random walks along  the network \citep*{noh2004}.  This is
the case of retrieving concepts since  the subject is not aware of his
full  semantic  network when  naming  them.   In  this kind  of  cases
reachability of nodes is  measured with the \textit{mean first passage
  time} (MFPT),  i.e. the averaged number  of steps needed  to visit a
node   $j$  for   the   first   time,  starting   from   a  node   $i$
\citep*{snell1959,catral2005}.  Given its  relevance in complex media,
this  paradigm has  been recently  revisited  in a  number of  studies
\citep*{noh2004,catral2005,condamin2007}.

While different derivations of  random-walkers have been recently used to infer
the   underlying   topological    properties   of   complex   networks
\citep*{costa2007,ramezanpour2007,gomezgardenes2008}, our aim was to evaluate how SRW (and
in  particular,   the  effect   of  different  levels   of  switching)
contributes  to the  exploration  of network  models  with well  known
topological  properties. Different models  which were  not necessarily
lexico-conceptual  architectures  were  explored  by  a  SRW  and  its
performance  was  measured by  the  MFPT (detailed  in  section  \ref{SRWmarkov}).  Going  back  to  the
cognitive paradigm,  retrieving plenty of  words in a  semantic verbal
fluency  test not  only depends  on the  number of  concepts  that the
subject  knows, but  also  on an  equilibrium  between the  underlying
semantic topology  that organizes those concepts and  the frequency of
switching  \citep*{troyer1997}.  For   example,  two  different  studies
\citep*{boringa2001,sepulcre2006} reported  that their respective groups
of healthy participants  produced $30.7 \pm 7.9$ and  $28.15 \pm 7.32$
animals  during  $90$ seconds. There  are two remarkable  aspects in these  figures. First,
participants  obviously knew many  more animals  than those  said and,
second, there is  a high heterogeneity in the  number of words. Hence,
even though all participants only  named a low fraction of the animals
they  knew,  some of  them  had much  more  success  than others  when
retrieving them.



\section{A Markov model of SRW}

As introduced in  the previous section, our approach  for a clustering
step consists of a walker unaware  of the full network moving from one
node  to any  of its  neighbors with  no preferential  gradients among
neighbors.  Such  exploration  task  was  modeled  by  the  well  known
\textit{random-walker} (RW).  Switching was implemented  as a mechanism
where  the  walker  moves   to  any  other  node  following  different
probabilistic  approaches. Summarizing  SRW  can be defined  as a  random-
walker with the capability of rendering random shifts.

\subsection{Markov Chains}
\label{markovchains}

A finite Markov chain is a special type of stochastic process which
can be  described as follows. Let
\begin{equation}
S=\{s_1,...,s_r\}
\end{equation}
be a finite  set whose members are the  \textit{states} of the system,
which we  label $s_1,...,s_r$. The process moves  through these states
in a sequence of \textit{steps}. If at any time it is in state $i$, it
moves  to  a  state  $j$  on  the next  step  with  some  probability,
$\Pi:S\times S\to {\cal M_{S\times  S}}$, where ${\cal M}_{S\times S}$
is  the set of  $S\times S$  matrices of non-negative entries where  the sum  of every  row is
1. These probabilities define a square, $r\times r$ matrix, $\Pi$:
\begin{equation}
\Pi\equiv[p_{ij}],
\end{equation}
which  we call  the \textit{matrix  of transition  probabilities}. The
importance of matrix theory to  Markov chains comes from the fact that
the $ij$th entry  of the $n$th power of  $\Pi$, $\Pi^n=[p_{ij}^{(n)}]$ represents the
probability  that the process  will be  in state  $j$ after  $n$ steps
considering that it  was started in state $i$. The  study of a general
Markov  chain can  be reduced  to the  study of  two special  types of
chains.   These  are  \textit{absorbing  chains}  and  \textit{ergodic
  chains} (also known as \textit{irreducible}). The former contain at
least  one absorbing  state, i.e.   a  state constituted  by a  proper
subset of  the whole  by which,  once entered it  cannot be  left, and
furthermore, which is reachable from every state in a finite number of
steps. The  latter are those chains  where is possible to  go from any
state to  any other state in a  finite number of steps  and are called
\textit{regular} chains when
\begin{equation*}
(\exists n<\infty): (\forall i,j\leq r)(\forall N> n)(p_{ij}^{(N)}> 0).
\end{equation*}
For regular chains, the $ij$th entry of $\Pi^n$ becomes essentially
independent of state $i$ as $n$ is larger.  In the case of regular chains, we can define
a stationary probability matrix \citep*{snell1959} $\Pi^{\infty}$ as:
\begin{equation}
\lim_{n\to\infty}\Pi^n=\Pi^{\infty}.
\end{equation}

Note that for non regular Markov processes this limit might not exist. For instance  $\Pi={\tiny \left( \begin{matrix} 0 & 1 \\ 1 & 0 \end{matrix}\right)}$.

The matrix $\Pi^{\infty}$ consists of a row probability vector $w$ which is repeated on each row. This vector $w$ can be obtained as the 
only probability vector satisfying $w=w\Pi$ \citep*{grinstead1952}. For the case of regular Markov processes obtained from random walks on graphs, this indicates that in the long run, the probability to be in a node is independent of the node where the process started.

\subsection{Graph Characterization}
\label{graphcharac}
This  section is  devoted to  the characterization  of  the underlying
object  over which  we apply  our algorithm  of exploration,  a graph.
Beyond its main features, we discuss the consequences of connectedness
in order to clearly define the frameworks over which the SRW algorithm
can be  defined. Finally, we  briefly define the graph models studied
numerically in section \ref{results}.
\begin{figure*}
\begin{center}
\includegraphics[width=15cm]{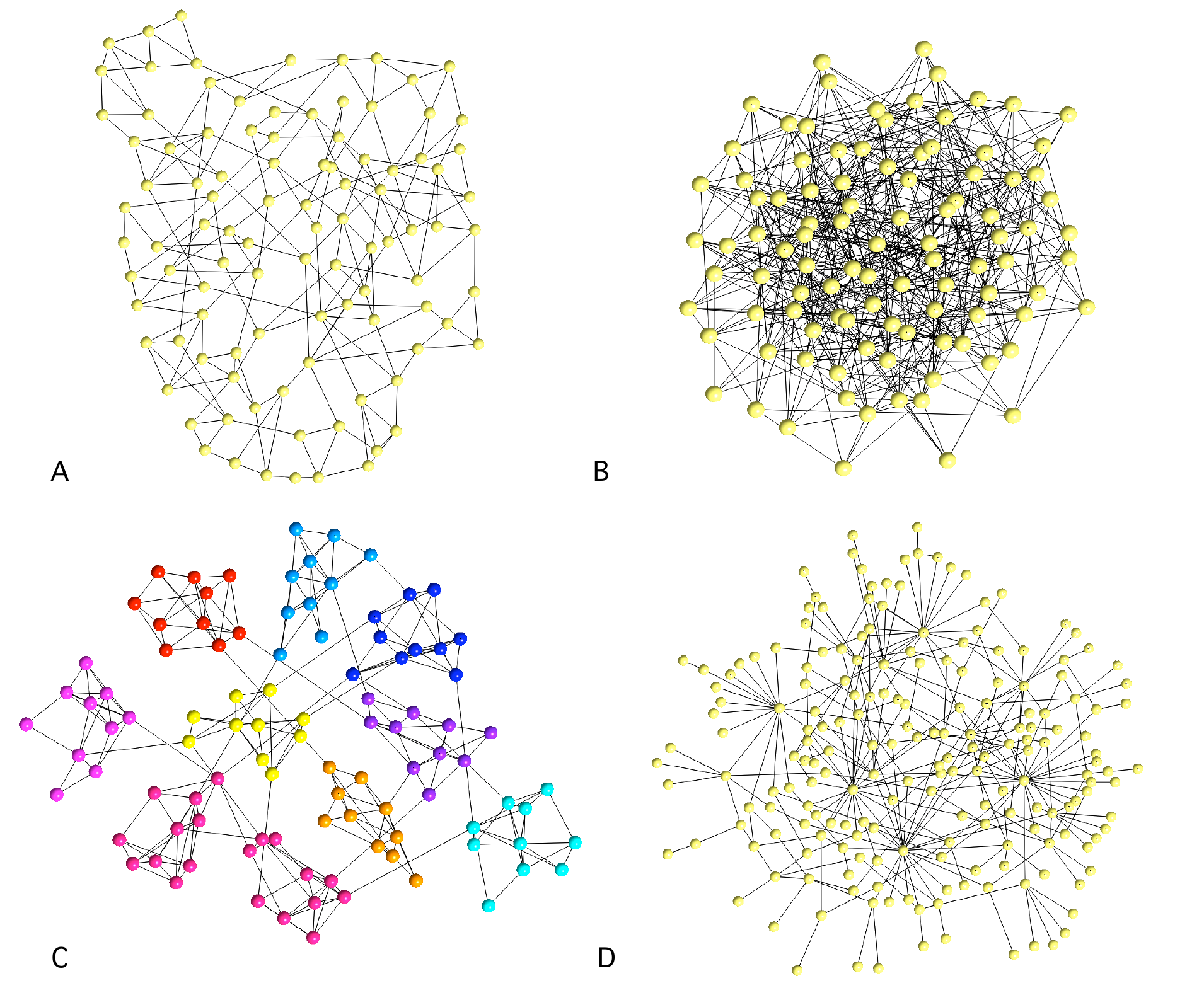}
\end{center}
\caption{\footnotesize{ Visualization of small examples ($|V|=100$) of the four network models analyzed here: (A)
    \textit{Small-world network}. (B)  \textit{Random
    Erd\"os-R\'enyi network}.  (C) \textit{Random-modular network}:  here a
    network is  partitioned into $10$ modules, each one
    connecting to each other with  a large probability, whereas a very
    small inter-module  probability is used. (D) \textit{Scale-free network}
    obtained by means of preferential attachment. See section \ref{graphcharac} for a detailed description of each network model \hfill}.}
\label{figSRWexample}
\end{figure*}

Let  us  suppose  that our  Markov  chain  is  defined by  some  graph
topology.   A  graph  ${\cal  G}$  is  defined  by  a  set  of  nodes,
$V\equiv\{v_1,...,v_n\}$,  and a set  of links  $\Gamma\equiv \{\{v_i,
v_j\},...,\{v_k,  v_l\}\}$,  being $\Gamma$ a subset of $V\times V$. In our  approach, the
graph is undirected and we  avoid the possibility that a node contains
auto-loops or  that two links are  connecting the same  nodes. The {\em
  size} of the graph is $|V|$, i.e. the cardinal of the set of vertices. Its {\em average connectivity}
is defined as:
\begin{equation}
\langle k\rangle\equiv\frac{2|\Gamma|}{|V|}.
\end{equation}
The topology of our graph is completely described by a symmetrical,
$|V|\times |V|$ matrix, $A({\cal G})=[a_{ij}]$, the so-called {\em adjacency
 matrix}, whose elements are defined as:
\begin{eqnarray}
a_{ij}=\left\{
\begin{array}{ll}
1\;\;\leftrightarrow \;\;\{v_i,v_j\}\in\Gamma\\
0\;\;{\rm otherwise}.
\end{array}
\right.
\end{eqnarray}
The connectivity  of the node $v_i$,  $k(v_i)$ is the  number of links
departing from $v_i$ and it  can be easily computed from the adjacency
matrix as:
\begin{equation}
k(v_i)=\sum_{j\leq |V|}a_{ij}.
\end{equation}
Following the characterization, we now define the degree distribution,
which is  understood as  the probability that  a randomly  chosen node
displays a given connectivity. In  this way, we define the elements of
such a probability distribution, $\{p\}$ as:
\begin{equation}
p_i=\frac{|(v_j\in V):(k(v_j)=i)|}{|V|}.
\end{equation}

The  above defined measures  are the  {\em identity  card} of  a given
graph ${\cal G}$.  One could think  that it is enough because our main
goal  is to describe  and characterize  an exploration  algorithm over
${\cal G}$.   However, specially  in the models  of random  graphs, we
cannot  be directly  sure that  our adjacency  matrix defines  a fully
connected graph, i.e,  that there exists, with probability  $1$ a path
from any node $v_i$ to any node $v_j$. In deterministic graphs, we can
solve  this problem  by  assuming, a  priori,  that our  combinatorial
object  is fully  connected. Furthermore,  we could  agree  that, when
performing  rewirings   at  random,   we  impose  the   condition  of
connectedness.   The  case  of  pure  random  graphs  is  a  bit  more
complicated.   Indeed, a  random  graph is  obtained  by a  stochastic
process of addition or removal of links \citep*{bollobas2001}. Thus, we
need a criteria to ensure that our graph is connected or, at least, to
work over  the most representative  component of the  obtained object.
Full connectedness is hard to ensure in a pure random graph. Instead,
what  we  can  find  is  a {\em  giant  connected  component},  $GCC$.
Informally speaking,  we can imagine an algorithm  spreading at random
links among  a set of predefined nodes, the so-called Erd\"os-R\'enyi
graph process.  The growing graph displays, at the beginning, a myriad
of small clusters of a few  nodes and, when we overcome some threshold
in the number  of links we spread at random,  a component much bigger
than the others emerges, i.e. the  $GCC$ \citep*{erdos1960}.  In this way, M.
Molloy  and B.   Reed  \citep*{molloy1995} demonstrated  that, given  a
random graph with degree distribution $\{p\}$, if
\begin{equation}
\sum_kk(k-2)p_k>0
\label{eqMolloy1995}
\end{equation}
then,  there   exists,  with  high  probability,   a  giant  connected
component. The  first condition  we need to  assume is thus,  that the
studied  graphs satisfy  inequality (\ref{eqMolloy1995}).   Beyond this
assumption, we  impose the following criteria when  studying our model
networks:
\begin{enumerate}
\item
In a deterministic graph (for example,  a chain or a lattice) where we
perform random re-wirings,  we do not allow re-wirings  that break the
graph.
\item
If a  graph is  the result of  an stochastic process,  the exploration
algorithm is defined  only over the $GCC$ (this could imply
the whole set of nodes).
\item
The adjacency  matrix is the adjacency  matrix of the  $GCC$.  We {\em
  remove}  the  nodes that,  in  the  beginning,  participated in  the
process of construction of ${\cal G}$ but fell outside the $GCC$.
\end{enumerate}
All  the  model  graphs  studied   in  this  work  satisfy  the  above
conditions.

In  order to enable  useful comparative  analysis, we  built different
networks,  all  of  them  with  $|V|=500$  nodes  and  $|\Gamma|=2000$
links. The  results were averaged  after $100$ instances per network model.  Let  us briefly
define the models we will study with our exploration algorithm.

{\bf Watts-Strogatz  Small-World Network}.   We built an  annulus with
$500$  nodes  in such  a  way  that every  node  is  connected to  $8$
different  nodes  (2000  undirected links)  \citep*{watts1998}.  Once  the  annulus  was
constructed, every  link suffered a random  re-wiring with connectivity
$p=0.05$.

{\bf Erd\"os R\'enyi  Graph}. Over a set fo $500$  nodes we spread at
random $2000$  links, avoiding duplication  and self-interaction.  It
can be  shown that the  obtained graph displayed a  binomial degree
distribution \citep*{erdos1960}:
\begin{equation}
p_k={|V|-1 \choose k}\pi^k(1-\pi)^{|V|-k-1},
\end{equation}
being $\pi$ the probability of two nodes being connected. Its value corresponds to
\begin{equation}
\pi=|\Gamma|{|V| \choose 2}^{-1}
\end{equation}

{\bf Random-Modular}. We built $10$ different components of $50$ nodes
and  $200$  links,  spread at  random (as explained for Erd\"os R\'enyi graphs)  among  the  $50$ nodes  of  every
component.  In this case,  we ensure connectedness of such components.
Once the ten  components are constructed, every link  suffers a random
rewiring  with a  node either  from the  same component  or  not, with
probability $p=0.05$.

{\bf  Preferential Attachment}.  We  provide a  seed of  $9$ connected
nodes.   Every new  node  was  connected to  $8$ of the  existing nodes  with
probability proportional  to the  connectivity of the  existing nodes,
i.e.,  suppose that, at  time $t$  a new  node $v_i$  comes in  to the
graph. At this  time step, the graph will  display an adjacency matrix
$A(t)$.
\begin{equation}
\mathbb{P}(a_{ij}(t)=1)=\frac{k(v_j)(t-1)}{\sum_{v_k\in {\cal A}_t}k(v_k)(t-1)},
\end{equation}
where
\begin{equation}
{\cal A}_t=\{(v_k: \exists l):(a_{kl}(t)>0)\}
\end{equation}
This  operation is repeated  in an  iterative fashion  (i.e., updating
$A$) $8$ times per node. It can be shown that, at the limit of a large
number of nodes the outcome  of this algorithm generates a graph whose
degree distribution is a power law \citep*{barabasi1999}:
\begin{equation}
p_k\propto k^{-\alpha},
\end{equation}
with $\alpha=3$. It is worth  noting that such an algorithm avoids the
possibility of unconnected components.

\subsection{Random walk over a graph as a Markov Process}
\label{SRWmarkov}

In this  framework, the transition  from node $i$  to $j$ is  just the
probability that a random-walker starts from some node $i$ and reaches
the  node $j$, after  some steps.  Consistently, the  probability that
being  in  $v_i$ we  reach  the  node $v_j$  in  a  single step  (i.e,
$p_{ij}$) is:
\begin{equation}
p_{ij}=\frac{a_{ij}}{k(v_i)}
\label{A/k}
\end{equation}
This is the general form for a Markov formalization of a random-walker
within a graph  defined by its adjacency matrix  $A$.  Throughout this
work  we   assume  that  our   graphs  define  {\em   regular}  Markov
processes (see section \ref{markovchains}). Under the above  definition of $\Pi$, regularity is assured
if  and only  if the  graph is  not bipartite  (i.e., it  contains, at
least,  one loop containing  an odd  number of  nodes). To see that bipartite graphs are not regular,  it is enough
to notice that for any pair of nodes $(v_i,v_j)$ there are only either odd or even paths joining them, but not both.
Hence if $p_{ij}^{(n)}\neq 0$ then  $p_{ij}^{(n+1)}=0$ and therefore the process cannot be regular.

Summarizing, despite that connectedness ensures  that the
process is ergodic,
\begin{equation*}
(\forall v_i, v_j\in {\cal G})(\exists n: p_{ij}^{(n)}\neq 0)
\end{equation*}
it does not ensure regularity and therefore the $\lim_{n\rightarrow \infty}\Pi^n$ might not exist. The existence  of an odd loop  breaks such parity  problem and enables
$\Pi^n$ to stabilize to  a specific matrix of stationary probabilities
when  $n\to\infty$.  Thus, we  must impose  another assumption  to our
studied graphs:  Our algorithm  works over non-bipartite  graphs which
satisfy the criteria imposed in section II B.  It is straightforward
to  observe that,  if the  assumption of  regularity holds,  the above
Markov process  has a  stationary state with  associated probabilities
proportional to the connectivity of the studied node \citep*{noh2004}:
\begin{equation}
p_{ij}^{(\infty)}=\frac{k(v_j)}{2|\Gamma|}.
\end{equation}
From now on,  we will refer to the transition  matrix above defined as
$\Pi^{cl}$, since it denotes the probabilities of the movements related
to clustering.

\subsection{Switcher-random-walker}
In  the retrieval  model  introduced here,  the  matrix of  transition
probabilities  $\Pi^{srw}$ is  a linear  combination of  the switching
transition  probabilities  $\Pi^{sw}$  and the  clustering  transition
probabilities $\Pi^{cl}$, as defined  in the above section. The \textit{Markov
process} is  a switcher-random-walker and  the \textit{states}
represent the location of such walker in the network.

\begin{figure*}
\begin{center}
\includegraphics[width=18cm]{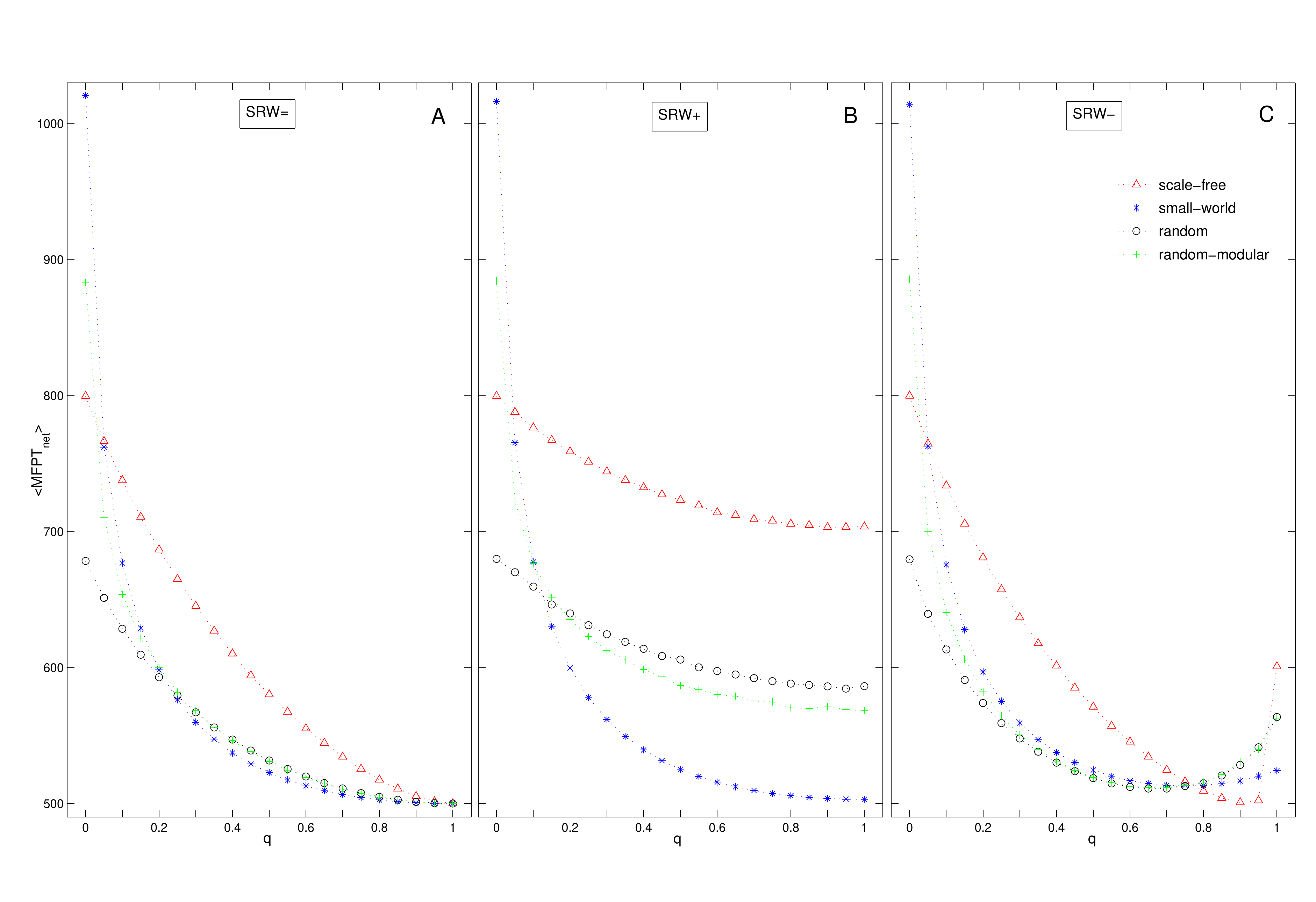}
\end{center}
\caption{\footnotesize{Exploration performance based on the $\langle MFPT \rangle_{\cal G}$ (see equation \ref{eqMFPTgraph}) on $4$ graph models  for the three Markovian variants of SRW (see equation \ref{eqdelta} for implementation details of each variant of switching).  Parameter $q$ stands for probability of switching (see equation \ref{eqpijsrwMatrix}).  
    (A) $SRW^=$, SRW that contains a uniformly distributed switching.
    (B) $SRW^+$, SRW that contains a switching with positive degree gradient.
    (C) $SRW^-$, SRW that contains a switching with negative degree gradient.\hfill}}
\label{figMFPTmosaic}
\end{figure*}

The matrix  $\Pi^{sw}=[p_{ij}^{sw}]$ is  ergodic and regular since all entries are strictly greater than zero,  and has
equal rows, i.e. constant columns.  The reason is that the probability
of reaching  a node $j$ through  switching is independent  of the source
node $i$. In this way, we could consider that we define a scalar field
$\lambda$ over the nodes of the graph:
\begin{equation}
p^{sw}_{ij}=\lambda_j.
\label{pswij}
\end{equation}
Consistently, 
\begin{equation}
\sum_{j\leq |V|}\lambda_j=1.
\end{equation}
We  can  define  this  field  in  many different  ways.  As  the  more
representative, we  revise several scalar  fields that can  provide us
interesting information about the process:
\begin{equation}
\lambda_j=\left\{
\begin{array}{ll}
\displaystyle\frac{1}{|V|} \\ \\
\displaystyle\frac{k(v_j)}{\displaystyle\sum_{i\leq |V|}k(v_i)}\\ \\
\displaystyle\frac{K-k(v_j)+1}{\displaystyle\sum_{i\leq |V|}k(v_i)}\\
\end{array}
\right.
\label{eqdelta}
\end{equation}
In the  first and most  simple case switching  to any other node  is a
random uniform process, and we refer to this process as {\em uniformly
  distributed  switching}.    The  second  case   corresponds  to  the
situation  where  the  probability  to  reach  a  given  node  through
switching  is proportional  to  its connectivity, which we call  {\em
  positive degree gradient switching.}   The last one assumes that $K$
is  $\max\{k(v_i)\}$  and  corresponds  to  the  situation  where  the
switcher jumps with more probability to weakly connected nodes, and we
refer to it as {\em negative degree gradient switching}. These three variants of switching
were studied  when combined with a random-walker within  the  above   graph  topologies  (see
fig.(\ref{figMFPTmosaic})). They were denoted by $SRW^=$, $SRW^+$ and $SRW^-$ respectively.


The matrix  $\Pi^{cl}=[p_{ij}^{cl}]$ defined  in the above  section is
ergodic and regular  but restricted to the transitions  allowed by the
adjacency  matrix $A$  of the  network of  study. We  modeled as
equi-probable the transitions among linked nodes of the network. Hence the probability of moving from a node $v_i$ to a
a node $v_j$ through clustering for a given graph ${\cal G}$ with an adjacency matrix $A_{\cal G}=[a_{ij}]$, is

\begin{equation}
p^{cl}_{ij}=\frac{a_{ij}}{k(v_i)}.
\label{pclustering}
\end{equation}

Thus, $\Pi^{srw}=[p_{ij}^{srw}]$ is defined as:
\begin{equation}
\label{eqpijsrwMatrix}
\Pi^{srw}=q\Pi^{sw}+(1-q)\Pi^{cl}\;\;(0\leq q\leq 1),
\end{equation}
where $q$ is the probability of switching. Consistently, the entries of $\Pi^{srw}$ are given by:
\begin{equation}
p_{ij}^{srw}=q p_{ij}^{sw} + (1-q) p_{ij}^{cl},\quad 0 \le q \le 1.
\label{eqpijsrw}
\end{equation}


We observed  that $\Pi^{srw}$
is also ergodic and  regular. This follows from the fact that $\Pi^{sw}$ has already all entries strictly greater than zero, and thus $\Pi^{srw}$ will have all entries greater
than zero for any $q>0$. For the case of $q=0$, $\Pi^{srw}$ is just $\Pi^{cl}$ which we assumed to be regular.

Among  other  interesting descriptive  random  variables  that can  be
evaluated  for regular chains,  the matrix  of the  \textit{mean first
  passage  time}   (MFPT)  is  a   matrix  $\langle  T\rangle=[\langle
  t_{ij}\rangle]$, crucial for  measuring the retrieval or exploratory
performance of  any stochastic strategy; the  MFPT needed to  go from a
node  $i$  to  a  node  $j$  is  denoted  by  $\langle  t_{ij}\rangle$
\citep*{noh2004}  and represents the  time (in  step units)  required to
reach state  $j$ for the  first time starting  from state $i$.   It is
important  to note  that  $\langle t_{ij}\rangle$  is not  necessarily
equal to $\langle  t_{ji}\rangle$, i.e. it might happen  that the time
required to  go from state $i$ to  state $j$ is different  to the time
required to go from state $j$ to state $i$.

In order to obtain the analytical expression of MFPT, we must define
first a fundamental matrix $Z$ \citep*{grinstead1952} which is
given by
\begin{equation}
Z=(\mathbb{I}-\Pi^{srw}+\Pi^{\infty}_{srw})^{-1},
\label{eqz}
\end{equation}
where
\begin{equation}
\Pi^{\infty}_{srw}=\lim_{n\to\infty}\left(\Pi^{srw}\right)^{n},
\end{equation}
and $\mathbb{I}$ is the identity matrix of size $|V|\times |V|$.

In this case  the entry $z_{ij}$ of $Z$  can be understood
as a measure of the  deviations of the $ij$th entry of $(\Pi^{srw})^n$
from their  limiting probabilities $w$, which, as commented in section \ref{markovchains}, is any of the equal rows of $\Pi^{\infty}_{srw}$. From  $Z$ and $w$ we can  obtain the
analytical derivation of $\langle T\rangle=[\langle t_{ij}\rangle]$ (for more details see  \citep*{grinstead1952}):


\begin{equation}
\langle t_{ij}\rangle = \frac{z_{jj}-z_{ij}}{w_j}
\end{equation}


Finally,  we denote  as  $\langle MFPT\rangle_{\cal  G}$ the  averaged
value  of all entries  $\langle t_{ij}\rangle$  for a  switcher random
walker exploring a network ${\cal G}$. Since $\langle T \rangle$ it is
not necessarily symmetrical, we must take into account all the entries
outside  the main diagonal.  The main  diagonal was not  taken into
account,  since it  represents the  returning  time, which  we do  not
consider as a part of the exploration of the net. Thus,
\begin{equation}
\langle MFPT\rangle_{\cal G}=\frac{1}{2{|V| \choose 2}}\sum_{i}\sum_{j\neq i}\langle t_{ij}\rangle.
\label{eqMFPTgraph}
\end{equation}
This measure  provides a  general evaluation of  how reachable  is, on
average, any node from any other node in a specific network using a switcher
random-walker. It  is interesting to  notice that such measure  has an
upper bound  which is precisely  the size of  the net.  Indeed,  let us
suppose we  have a  clique of  size $m$, i.e.,  a graph,  ${\cal G}(V,
\Gamma)$, where $|V|$ equals $m$ and every node $v_i$ is connected to itself and to all $m-1$
remaining nodes.  It corresponds to  the case where the probability of
switching  is $1$. Let  $X$ be  a random  variable whose  outcomes are
$v_j$ such that, $\forall v_j\in V$:
\begin{equation}
\mathbb{P}(X=v_j)=\frac{1}{m}.
\end{equation}
We  define  a stochastic  process,  namely,  the  realizations of  $X$
through  different  time  steps,  $X(1),X(2),...,X(t)$ Let  us  define
another random variable, $Y$, namely the number of realizations of $X$
needed to ensure  that there has been one realization  of $X$ equal to
$v_j$:
\begin{equation}
Y=\min_t\{X(t)=v_j\}
\end{equation}
 Clearly, and  due to  the symmetry of  our experiment, all  the nodes
 behave in the same way. Furthermore,
\begin{equation}
\langle Y\rangle=m
\end{equation}
i.e., we need, in average $m$  realizations of $X$ in order to obtain,
at least, one realization $X=v_j$, $\forall v_j\in V$. We observe that
the above random experiment is exactly a random switching over a graph
containing  $m$ nodes,  and that  $\langle Y\rangle$  is  the $\langle
MFPT\rangle$  of this  process.  Let  us  suppose we  have a  $\langle
MFPT\rangle<m$. This implies that, in average
\begin{equation}
(\forall v_j)\mathbb{P}(X=v_j)>\frac{1}{m}
\end{equation}
which is a  contradiction, since the graph has $m$  nodes. Thus, for a
given graph ${\cal G}(V,\Gamma)$:
\begin{equation}
\langle MFPT\rangle_{\cal G}\geq |V|.
\end{equation}
This value represents  an horizontal asymptote in the  model of SRW as
$q$ increases, and it is clearly defined in our model experiments (see
figure \ref{figMFPTmosaic}).
 
\section{Results and discussion}
\label{results}

Our  main  result  was  that  SRW exploration,  a  cognitive  inspired
strategy  that  combines  random-walking  with  switching  for  random
exploration of networks, decreased the $\langle MFPT \rangle$ of all  models for all SRW
variants. This means  that, on average, the number  of steps needed to
travel  between every  pair of  nodes decreases  and thus  the overall
exploration abilities of a SRW within the networks improves respect to
RW.

Regarding    $SRW^{=}$   (Fig.    \ref{figMFPTmosaic}A),   exploration
performance  of  random-modular   and  small-world  networks  severely
improves, overtaking  scale-free at $q=0.1$. Moreover,  at $q=0.3$ all
the  networks  but  scale-free  converged,  leading  to  a  remarkable
scenario  where modularity  and high  clustering coefficients  are not
topological handicaps for an efficient information retrieval.

Switching  in   $SRW^{+}$  severely  improves  $\langle MFPT \rangle$   in  modular  and
small-world  networks  while hardly  decreases  it  in scale-free  and
random. The  reason is that a  random-walker on both  kind of networks
already   shows   a  gradient   to   visit   highly  connected   nodes
\citep*{noh2004}, and a  positive-degree switching supported rather than
compensated    this    effect    due    to    redundancy    on    hubs
\ref{figMFPTmosaic}B).

In $SRW^{-}$,  intermediate values  of q (around  $0.6$ for
all but scale-free models) showed optimal performance with a similar effect to the one produced by $SRW^{=}$.  However, it only  partially succeeded in compensating   the   already  commented   natural   RW  gradient   for
hubs. (Fig. \ref{figMFPTmosaic}C). Interestingly, those $q$ values close to $1$ produced an
inverse situation where hubs are so unlikely to be reached that the overall exploration
performance decreased for all the models but dramatically for scale-free model, where the 
degree heterogeneity is specially high. On the contrary, small-world model showed a very similar 
performance when explored by any of the three SRW variants. The reason is that in this model, the degree distribution 
is very homogeneous, and thus different degree gradients of switching produced very little differences.

The approximate convergence of the exploration efficiency (for most of
the topologies when  using $SRW^=$ or $SRW^-$ with  a moderate switching
rate) allows  a system  to organize information  or to  evolve without
compromising  exploration  and retrieval  efficiency.  In this  sense,
semantic memory might be  organizing information in a strongly modular
or locally clustered way without compromising retrieval performance of
concepts. In a  more general perspective, the addition  of a switching
mechanism and its interaction  with random-walker dynamics opens a new
framework to  understand processes related to  information storage and
retrieval. Indeed,  switching not only  mitigates exploration deficits
of  certain   network  topologies  but  also   might  provide  certain
robustness to  the system. For  instance, the rewired links  (known as
short-cuts)  in   both  small-world  and   random-modular  models  are
contributing  to  facilitate  access   to  different  regions  of  the
network. Those  short-cuts might compensate a  switching impairment or
dysfunction and  vice versa, i.e.  switching would  ensure an accurate
exploration of the network even though a targeted attack removed those
short-cuts permanently.

Similar mechanisms to  switching have been observed in  the context of
information   networks.   In   particular,  the   iterative  algorithm
\textit{PageRank}  estimates   a  probability  distribution   used  to
represent the likelihood that a person randomly clicking on links will
arrive  at any  particular page  for a  hyper-linked set  of documents
(e.g. the  world-wide-web) \citep*{brin1998}. The user  is supposed to
be a \textit{random-surfer} who begins  at a random web page and keeps
clicking on links but  never hitting back. The \textit{damping-factor}
is an  additional item that  includes the fact  that the user  can get
bored and start  on another random page. The  combination of these two
processes is used by the Google Internet search engine to estimate the
relevance of  different links (PageRank  values). Interestingly, while
the  objective (rank  link  targets) and  the framework  (hyper-linked
documents,   i.e.   directed   graphs)   are   not   the   same,   the
cognitive-inspired SRW  described here and  PageRank algorithm combine
random-walks  restricted  to  a  topology  with  an  extra-topological
mechanism in order to evaluate tasks in complex networks.


The model proposed here could  have implications in other systems that
usually  have  a  conflict   between  organization  and  retrieval  or
spreading efficiency.  It will be  object of further studies  in other
phenomena   unrelated  to  cognitive   processes  such   as  infection
epidemiology, information spreading or energy landscapes.

\noindent {\bf Acknowledgments} \smallskip

\noindent JG is a fellow of  the Government of Navarra. IM is a fellow
of  the Caja  Madrid  Foundation.  This work  was  supported by  James
McDonnell Foundation to BCM, MEC of Spain BFM2006-03036 to SAT and the
European  Commission  (NEST-Pathfinder:  ComplexDis  contract  number:
043241) to  PV. Thanks to Ricard  V. Sol{\'e} for  his useful comments
and support in the design of figures.

\end{document}